\documentclass[conference]{IEEEtran}
\IEEEoverridecommandlockouts
\usepackage{cite}
\usepackage{amsmath,amssymb,amsfonts}
\usepackage{algorithmic}
\usepackage{graphicx}
\usepackage{textcomp}
\usepackage{xcolor}

\usepackage{multirow}
\usepackage{booktabs}
\usepackage{diagbox}
\usepackage{subfigure}
\usepackage{afterpage}
\usepackage{float}
\usepackage{hyperref}
\usepackage{array}
\usepackage{lipsum}
\usepackage{url}

\def\BibTeX{{\rm B\kern-.05em{\sc i\kern-.025em b}\kern-.08em
    T\kern-.1667em\lower.7ex\hbox{E}\kern-.125emX}}

\begin{document}

\title{GANSER: A Self-supervised Data Augmentation Framework for EEG-based Emotion Recognition 
\thanks{This work was supported by the Natural Science Foundation of Guangdong Province (No. 2019A1515011181), the Science and Technology Innovation Commission of Shenzhen under Grant (No. JCYJ20190808162613130), the National Natural Science Foundation of China (No. 62002230), and the Shenzhen high-level talents program.}}

\author{\IEEEauthorblockN{Zhi Zhang}
\IEEEauthorblockA{\textit{College of Computer Science and Software Engineering,} \\
\textit{Shenzhen University}\\
Shenzhen, China \\
zhangzhi2018@email.szu.edu.cn}

\\
\IEEEauthorblockN{Yan Liu}
\IEEEauthorblockA{\textit{Department of Computing,} \\
\textit{The Hong Kong Polytechnic University
Hong Kong}\\
Hong Kong, China \\
csyliu@comp.polyu.edu.hk}

\and
\IEEEauthorblockN{Sheng-hua Zhong$^{\ast}$\thanks{$^{\ast}$S.-h. Zhong is the corresponding author of this paper.}}
\IEEEauthorblockA{\textit{College of Computer Science and Software Engineering,} \\
\textit{Shenzhen University}\\
Shenzhen, China \\
csshzhong@szu.edu.cn}
}
\maketitle

\begin{abstract}
The data scarcity problem in Electroencephalography (EEG) based affective computing results into difficulty in building an effective model with high accuracy and stability using machine learning algorithms especially deep learning models. Data augmentation has recently achieved considerable performance improvement for deep learning models—increased accuracy, stability, and reduced over-fitting. In this paper, we propose a novel data augmentation framework, namely Generative Adversarial Network-based Self-supervised Data Augmentation (GANSER). As the first to combine adversarial training with self-supervised learning for EEG-based emotion recognition, the proposed framework can generate high-quality and high-diversity simulated EEG samples. In particular, we utilize adversarial training to learn an EEG generator and force the generated EEG signals to approximate the distribution of real samples, ensuring the quality of augmented samples. A transformation function is employed to mask parts of EEG signals and force the generator to synthesize potential EEG signals based on the remaining parts, to produce a wide variety of samples. The masking possibility during transformation is introduced as prior knowledge to guide to extract distinguishable features for simulated EEG signals and generalize the classifier to the augmented sample space. Finally, extensive experiments demonstrate our proposed method can help emotion recognition for performance gain and achieve state-of-the-art results.
\end{abstract}

\begin{IEEEkeywords}
EEG-based emotion recognition, data augmentation, generative adversarial networks
\end{IEEEkeywords}


\section{Introduction}
Emotions are manifest in each action of our daily life behaviors. Understanding emotions is one of the most important aspects of human development and growth, and, therefore, it is an important tile for the emulation of human intelligence \cite{cambria2017affective}. Thus, affective computing and automatic emotion recognition are key for AI advancement \cite{minsky2007emotion} and all the research fields that stem from them. Electroencephalography (EEG) measures oscillations in the brain, which reflect the synchronized activity of neurons. It is thought that the changes in these oscillations are correlated with the cognitive process, and can be used to reveal important information about human emotional states. As a kind of physiological signal, EEG has the advantage of being difficult to hide or disguise. Compared with other physiological signals, it has excellent time resolution, which is similar to the nuanced changes of emotional states in time scale. Owing to the rapid development of noninvasive, easy-to-use and inexpensive recording devices, EEG-based emotion recognition has received an increasing amount of attention in both research and applications \cite{zhong2020eeg}.

Nevertheless, EEG also subjects to several limitations. First, as an aggregate signal from the activity of millions of neurons, EEG suffers from a low signal-to-noise ratio (SNR) \cite{roy2019deep}. Second, EEG is generally recorded using tens to hundreds of electrodes simultaneously, and the sampling time usually exceeds a few seconds in each trial. Thus, the original feature dimension of an EEG sample is not low. However, in a typical dataset for cognitive neuroscience tasks, it usually contains only some hundred to a few thousand samples (i.e., experimental trials). It leads to a very low initial ratio of samples to features. Third, EEG is a non-stationary signal and its statistics varying over time. The inherent variabilities in brain anatomy, head size, and dynamics across trails/subjects considerably limit the generalizability of EEG analyses across subjects, and even across trials within a single subject performing a single task. The second limitation brings huge difficulties to the use of machine learning models, while the other two exacerbate this difficulty.

In the past few years, deep learning methods have achieved breakthrough performance for EEG-based emotion recognition. Unfortunately, deep learning models are typically very complex, i.e., have many free parameters (or degrees of freedom) to fit \cite{lashgari2020data}. Thus, if we lack enough data to train, considered the case of EEG-based emotion recognition that has a low initial ratio of samples to features, training such deep learning models risks overfitting those models to specific quirks of the training set. It also severely limits the generalizability of these models.

Data augmentation is considered as one of the effective technologies for solving the data scarcity problem. It is usually a process of generating the new realistic-like data by applying a transformation to the real data \cite{fawzi2016adaptive}. It also holds the promise to increase the accuracy and stability of the classification or regression. To overcome the data scarcity problem, in this paper, we propose a Generative Adversarial Network-based Self-supervised Data Augmentation (GANSER) framework for EEG-based emotion recognition. The proposed framework comprises two networks, including the Adversarial Augmentation Network (AAN) and Multi-factor Training Network (MTN). In the AAN, we propose a Masking Transformation operation to mask parts of EEG signals and then force the proposed Generative Adversarial Network (GAN) seeking to synthesize potential EEG signals based on the remaining parts. Here, the UNet, Channel Masking operation and STNet are employed to model the spatio-temporal features of EEG signals while adversarial training forces the generated EEG signals to approximate the distribution of real ones, ensuring the quality of simulated EEG signals. Next, in the MTN, the simulated EEG signals are utilized for training the emotion recognition models as augmented samples. The Multi-factor Self-supervised Learning loss is proposed to introduce the masking possibility as prior knowledge to guide the model extracting distinguishable features for simulated EEG signals and generalize the classifier to the augmented sample space.


In summary, the contributions of this paper can be highlighted as follows. (i) This paper proposes GANSER, permitting to tackle the bottleneck of data scarcity for EEG-based emotion recognition. (ii) In this paper, we are the first to combine adversarial training with self-supervised learning to synthesize real-like diverse EEG signals, and utilize the augmented EEG samples to self-supervise emotion recognition learning. On the one hand, adversarial training is designed to learn an EEG generator and force the synthesized EEG signals fitting the real distribution to augment real-like high-quality samples. On the other hand, a transformation function is employed to mask parts of EEG signals and force the generator seeking to synthesize diverse augmented samples different from given samples. The prior knowledge during transformation is utilized to guide the self-supervised learning upon augmented samples. (iii) Extensive experiments are carried out, and the results show that our proposed deep framework significantly outperforms the existing state-of-the-art methods. Finally, we adopt a quantitative assessment approach for EEG analysis to evaluate the quality and diversity of the augmented samples, and visualization results are provided for qualitative analysis.


\section{Related Work}
\subsection{EEG-based Emotion Recognition}

Recent years, boosted by the success of deep neural networks, deep learning-based emotion recognition \cite{8320798,9154557,9204431,8089737,9321519} has received an increasing amount of attention in both research and applications and these studies seek to explore end-to-end methods to tackle the EEG-based emotion recognition task.

In detail, deep neural networks such as recurrent neural networks (RNNs), 2D/3D convolutional neural networks (CNNs), or both were employed for feature extraction and classification. In 2016, Zhang \textit{et al.} \cite{zhang2018spatial} proposed a spatial-temporal recurrent neural network (STRNN) to investigate both spatial and temporal dependencies of EEG signals and achieve the state-of-the-art. In 2018, Li \textit{et al.} further \cite{li2016emotion} proposed a hybrid deep learning structure based on a CNN and an RNN for emotion recognition based on multi-channel EEG signals. The proposed method showed effectiveness in the trial-level emotion recognition task. In the same year, Salama \textit{et al.} \cite{salama2018eeg} employed 3D CNNs to classify human emotion. To feed an EEG signal into inputs of a 3D CNN, they divided the 2D shape (channel$\times$time) of EEG data into 6-s segments and stacked them along the third axis. In the following study, Moon \textit{et al.} \cite{moon2018convolutional} pointed out the limitation that only signals or features from individual electrodes are considered and employed brain connectivity features to account for synchronous activations of different brain regions. In 2020, Moon \textit{et al.} \cite{moon2020madenet} further improved their research and introduced three different types of connectivity measures to model brain connectivity with a CNN. Furthermore, two data-driven methods are proposed to construct the connectivity matrix and maximize classification performance. Luo \textit{et al.} \cite{luo2018wgan} found that EEG signals significantly varied depending on the individual and imposed difficulty in achieving satisfactory classification performance. They proposed a Wasserstein GAN-based framework to solve the domain shift problem by narrowing down the gap between the probability distribution of different subjects. Recently, Moon \textit{et al.} \cite{moon2020madenet} proposed to learn feature space mapping and perform individuality detachment to reduce subject-related information from EEG signals. The proposed method can effectively discard the subject-related information and perform well on emotion recognition tasks.

\begin{figure*}[]
  \begin{center}
      \includegraphics[width=\linewidth]{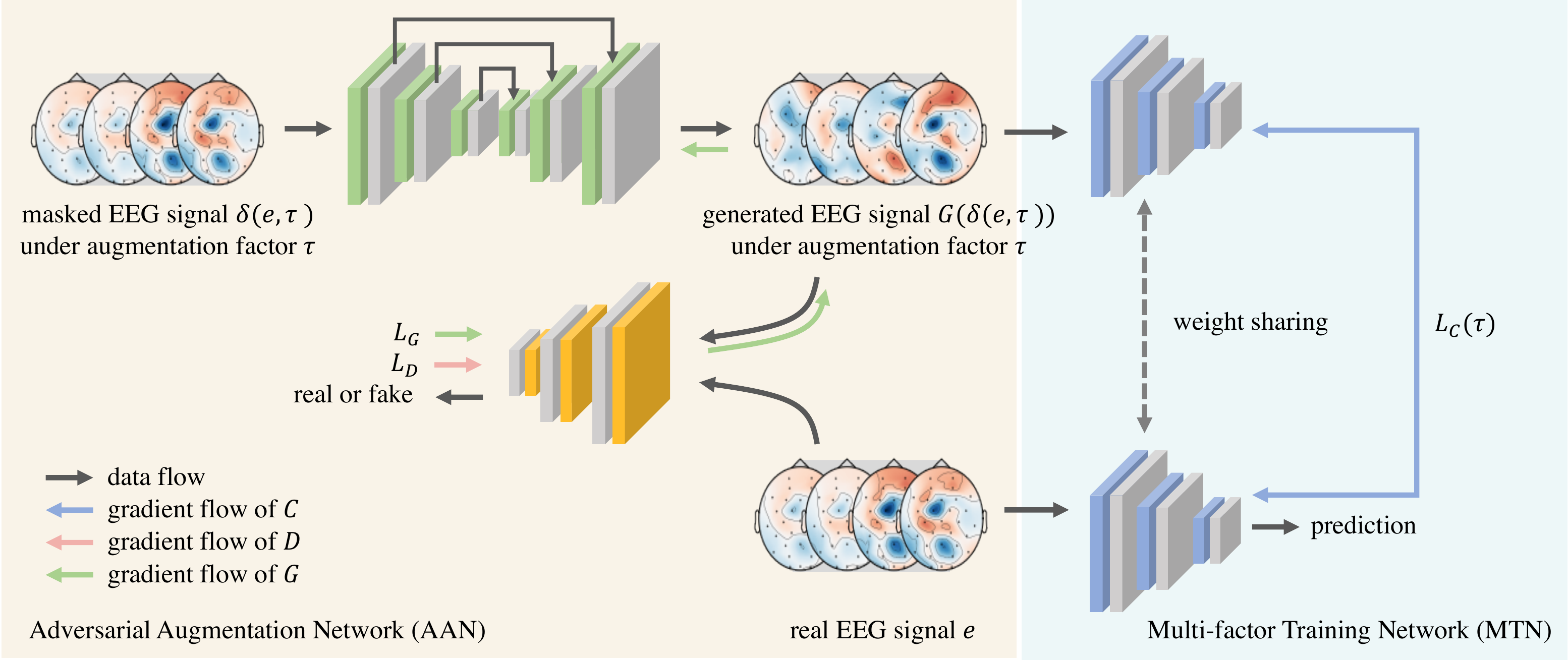}
  \end{center}
  \caption{Generative Adversarial Network-based Self-supervised Emotion Recognition (GANSER).}
  \label{framework}
\end{figure*}

\subsection{Data Augmentation for EEG-based Emotion Recognition}
In the field of image processing and computer vision, there are two simple and direct ways to augment data: geometric transformation and noise addition. Krell and Su \textit{et al.} suggested using rotational distortions, which were similar to affine/rotational distortions of images, to generate augmented EEG data \cite{krell2017rotational}. Relying on relevant combinations and distortions of the original trials, Lotte proposed three methods to obtain artificial EEG trials \cite{lotte2015signal}. Different from geometric transformations, Wang \textit{et al.} generated new features by adding Gaussian noises with different standard deviations to the original EEG feature and applied several deep learning models to verify the effect \cite{wang2018data}. Other data augmentation methods include sliding window, sampling, the Fourier transform, and recombination of segmentation \cite{lashgari2020data}. All of the abovementioned methods reported that the data scarcity problem had been alleviated, and the performance of the classifiers was improved \cite{luo2020data}.

Recently, Generative Adversarial Networks (GANs) have revealed their potential in generating EEG signals that mimic real ones, utilized in the emotion recognition task \cite{luo2018eeg,luo2019gan,luo2020data} and a wide variety of applications \cite{corley2018deep,ma2019depersonalized,aznan2019simulating,luo2020eeg,roy2020mieeg}. A conditional version of the Wasserstein Generative Adversarial Network (WGAN) was used to augment EEG data for emotion recognition in \cite{luo2018eeg}. They tried different sizes for the augmented data, and they found that doubling the data led to the highest performance increment comparing to other sizes. An SVM classifier trained on the augmented dataset improved 2.97\% for the SEED dataset from 83.99\% to 86.96\%. Luo \textit{et al.} proposed to use a conditional Boundary Equilibrium GAN (cBEGAN) to generate artificial differential entropy features of original EEG data, eye movement data and their concatenations for multi-modal emotion recognition. The main advantage of it is that the proposed GAN has good stability and a very quick convergence speed \cite{luo2019gan}. Luo \textit{et al.} proposed three methods for augmenting EEG training data to enhance the performance of emotion recognition models, including conditional Wasserstein GAN, selective variational autoencoder, and selective WGAN \cite{luo2020data}. They trained SVM and deep neural networks on original and augmented training datasets. The experimental results showed that the augmented training datasets enhance the performance of EEG-based emotion recognition.

Though lots of efforts have been made, the research on data augmentation for emotion recognition is far from close. For example, while a human can easily decide whether an augmented dataset, e.g., of cats or other images, still resembles the original class, the same is not true of augmented signals. How to measure the quality and diversity of augmented samples and synthesize high-quality and diverse augmented samples deserves further exploration.



\section{Proposed Method}
\label{sec:method}
\subsection{Overall Framework}

This paper designs a Generative Adversarial Network-based Self-supervised Emotion Recognition (GANSER) framework for EEG-based emotion recognition. Illustrated in Fig. \ref{framework}, the proposed framework comprises two networks, the Adversarial Augmentation Network (AAN), and the Multi-factor Training Network (MTN). Taking real EEG samples as input, the AAN is first designed to synthesize high-quality and diverse augmented EEG samples. Then, the EEG-based emotion recognition classifier can be learned on the augmented EEG samples and finish the self-supervised learning under the guidance of the proposed MTN. In the remainder of this section, we will detail the network architecture of AAN and MTN proposed in this paper.

\subsection{Adversarial Augmentation Network}


In the AAN, the Masking Transformation operation is first proposed to cut out part of the data points of the given EEG randomly. Then, a GAN is required to synthesize EEG signals fitting the real data distribution based on the remained EEG signals. By restoring the missing data points of EEG signals, the proposed GAN would be able to recognize the feature distribution of source EEG signals and introduce new data points to generate new EEG signals.

To be specific, given a 32 channel EEG signal with one-second length (sampled to 128 Hz) with the size of $32 \times 128$, we first follow the pre-processing of existing work \cite{yang2018emotion} to apply baseline removal, measuring the differences between baseline signals and the given signal. Then, the results of 32 channels are transformed into $9 \times 9$ maps according to the electrodes' location based on the international 10-20 system \cite{yang2018emotion}. As a result, the given EEG signal can be denoted by $e\in\mathbb{R}^{128\times 9\times 9}$.

Then, as we described before, we design the Masking Transformation operation to cut out partial signal values of the given $e$ and force the following GAN to restore the missing parts fitting the remained information to involve potential real-like samples different from the input EEG signal. In detail, we first randomly sample a matrix $r\in\mathbb{R}^{128\times 9\times 9}$ of the same size as $e$ with uniform distribution $U\sim[0,1)$ and utilize $r$ as the probability matrix representing probabilities of signal values being masked. Then, the parameter $\tau$ is sampled from the uniform distribution $U\sim[\tau_{min}, \tau_{max}]$ as the threshold to determine which data point should be masked. In this way, the obtained EEG signals $\delta(e, \tau)$ transformed from $e$ based on the threshold $\tau$ can be defined by:

\begin{equation}
  \label{eq:masked_eeg}
  \delta(e_{ijk}, \tau)=\left\{\begin{matrix}
    e_{ijk}&,r_{ijk} > \tau\\ 
    0      &,r_{ijk} \leq \tau
  \end{matrix}\right.
\end{equation}

Here, a large $\tau$ means random masking ignores more parts of the signal values of source EEG samples. In this case, the feature distribution of source EEG signals is hard to preserve due to the limited remained signal values. As a result, we can avoid learning an identity mapping and produce simulated EEG signals different from original signals, ensuring the diversity of augmented samples. Conversely, the generated EEG signal can be similar to the source EEG signal forced to fit the distribution of given signals. Thus, in this paper, we utilize $\tau$ as an augmentation factor to represent the augmented sample's diversity and difference from the original sample.

Then, based on $\delta(e, \tau)$, we design a GAN to synthesize simulated EEG samples and ensure the generated EEG signals fit the feature distribution of real samples. Unlike the Masking Transformation operation, focusing on involving the diversity for synthesized EEG samples at the signal value level, the proposed GAN is responsible for learning the distribution of realistic EEG signals at the feature level. In this way, the generated augmented samples are further forced to preserve the natural features of real samples. Finally, realistic and diverse samples can lead to better classification performance for emotion recognition.

The designed GAN is composed of two networks, i.e., a generator $G$ and a discriminator $D$, optimized to minimize a two-player min-max game. Here, the generator $G$ is trained to generate the simulated EEG sample $G(\delta(e, \tau))$ taking the EEG signal $e$ as input. The discriminator $D$ is required to distinguish whether the given EEG signals are simulated or real, while $G$ learns to fool the discriminator and try to make simulated samples close to real ones. Due to the instability problems of the traditional training procedure of GANs, different from previous GANs proposed for emotion recognition, we utilize a modified version of Wasserstein GAN Gradient Penalty, i.e., WGAN-GP \cite{NIPS2017_7159}, for combining adversarial supervision and the random masking augmentation strategy. Then, the loss function of $G$ can be formulated as Eq. (\ref{eq:g_loss}):

\begin{equation}
  \label{eq:g_loss}
  L_{G}=-\mathbb{E}_{e \sim \mathbb{P}_{e}}\left[D\left(G(\delta(e, \tau))\right)\right]
\end{equation}
where $\mathbb{P}_{e}$ denotes the distribution of the given real EEG signals and $e$ represents an EEG signal sample from it. Meanwhile, the goal of $D$ is to minimize the loss function illustrated in Eq. (\ref{eq:d_loss}):
\begin{equation}
  \label{eq:d_loss}
  \begin{aligned}
  L_{D}=&\mathbb{E}_{e \sim \mathbb{P}_{e}}\left[D\left(G(\delta(e, \tau))\right)\right]-\mathbb{E}_{e \sim \mathbb{P}_{e}}\left[D(\delta(e, \tau))\right] \\
  &+\lambda_p \mathbb{E}_{\hat{e} \sim \mathbb{P}_{\hat{e}}}\left[\left(\left\|\nabla D(\hat{e})\right\|_{2}-1\right)^{2}\right]
\end{aligned}
\end{equation}
where $\mathbb{P}_{\hat{e}}$ is defined sampling uniformly along straight lines between pairs of points sampled from the data distribution $\mathbb{P}_{e}$ and the generator distribution among $G(\delta(e, \tau))$. The gradient of the discriminator $D$ is denoted by $\nabla D(\hat{e})$, and $\lambda_p$ is a hyperparameter presenting the weight of the penalty term. In this way, the Wasserstein distance is used to compare the distributions of the generated samples and real samples, where the Lipschitz-continuous map ensures the property of a uniformly continuous distribution. This design can limit the normal of the derivation from growing too large \cite{NIPS2017_7159}. Utilizing Eq. (\ref{eq:g_loss}) and Eq. (\ref{eq:d_loss}), $G$ and $D$ are optimized in turn. By optimizing the adversarial loss, $D$ is able to distinguish real distribution from simulated distribution, while $G$ improves the ability to construct samples closer to real EEG signals.

Regarding the network architectures of $G$ and $D$, unlike the existing work focusing on generating feature representation or single-channel EEG signals, this paper aims to tackle the data augmentation problem for general emotion recognition methods and synthesize high-resolution EEG samples.

\begin{figure}[]
  \begin{center}
      \includegraphics[width=\linewidth]{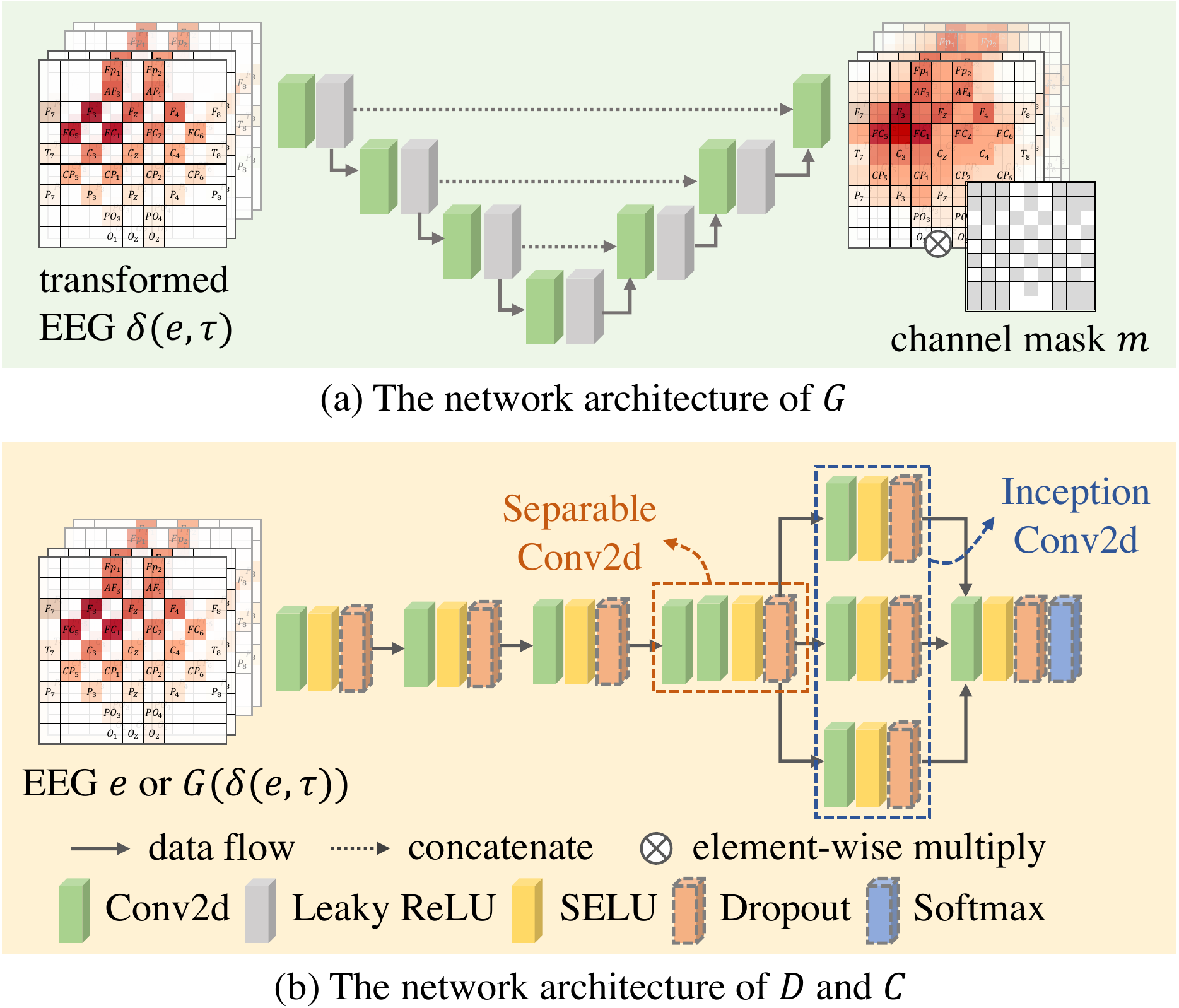}
  \end{center}
  \caption{Network architectures of proposed generator $G$, discriminator $D$, and classifier $C$.}
  \label{network}
  \vspace{-4pt}
\end{figure}

For the generator $G$, we aim to synthesize missing signal values of given EEG signals to augment new EEG samples. In this case, one of the natural ideas is to utilize an auto-encoder to reconstruct EEG signals directly in a down-sample and up-sample fashion. However, EEG signals contain abundant details reflected by time-series variance. It is challenging for existing auto-encoder-based networks to generate low-distorted EEG signals, because down-sampling of auto-encoder applied to high-resolution time sequences can lead to details missing and the reconstructed signals could be smoothed. To tackle this bottleneck, in this paper, we integrate an adaption version of UNet on EEG signals as the proposed generator for EEG signal synchronizing.

As shown in Fig. \ref{network}(a), the proposed UNet consists of an encoder, a decoder, and skip connections. The encoder takes the EEG representation as the input and utilizes four two-dimensional convolutional layers followed by LeakyReLU as an activation function to down-sample the EEG signals and extract feature maps. Based on the deep stack of convolutional layers, spatio-temporal patterns of original EEG signals are captured from detail to abstract. Then, in the decoder, three de-convolutional layers are applied to up-sample feature maps to high spatio-temporal resolution by synthesizing the missing signal values to generate new EEG samples based on extracted features. Here, skip connections are designed between the convolutional and symmetric de-convolutional layers to fuse shallow feature maps to favor de-convolutional layers to supplement high-resolution details.

Further, this paper finds EEG signals are sparse on the spatial dimension due to limited numbers of electrodes placed on the cap, which brings challenges for UNet to synthesize real-like EEG signals. To be specific, given a 32 channel EEG signal, we transform 32 channels into $9 \times 9$ maps according to the location of electrodes. In the locations where electrodes do not exist, signal values are unknown or unmeasured, and the signal values are represented as zeros. Forcing the UNet to fit unmeasured signal values in the locations where electrodes do not exist, signal values to 0.0, the generator is required to predict mutated low signal values into original spatially dense and continuous signals, which is at variance with objective reality and inevitably affects the modeling of measured signal values to fit the real distribution of original EEG signals. Thus, this paper first introduces the Channel Masking operation to improve the ability of UNet for synthesizing EEG signals. For the first step, we propose to build the channel mask $m$, a prior binary mask with the size of $9 \times 9$, and set the values where electrodes exist and signals are measured to one, while defining the other locations to zero. Then, we apply element-wise multiply between the designed prior mask and the output of UNet, i.e., the synthesized EEG signals, to artificially reset the signal values where electrodes do not exist to zero. In this way, the proposed generator only needs to focus on fitting the signal values where electrodes exist and neglecting the unreal mutation of EEG signal values caused by the nonexistent electrodes.

As the last part, illustrated in Fig. \ref{network}(b), we design a novel network architecture, STNet, to analyze the complex spatio-temporal features of EEG signals, and utilize STNet as the discriminator. In detail, the designed STNet comprises three two-dimensional convolutional layers, a separable convolutional layer and an Inception block. For the first step, input EEG signals are analyzed by two-dimensional convolutional layers to extract feature maps from signal values of each electrode and their spatio-temporal neighbors to summarize high-level features. Then, due to the fact that the recognition of specific emotions is only related to local patterns of spatial features or temporal features, we introduce a separable convolutional layer \cite{chollet2017xception} to decouple the modeling of spatio-temporal information. Here, the utilized separable convolutional layer containing a depth-wise convolutional layer and a point-wise convolution layer to capture spatial correlation and temporal correlations of extracted feature maps, respectively. Recognizing the pattern of emotions requires analyzing EEG signals at different spatial scales, and thus we further introduce an Inception block \cite{szegedy2015going} containing three types of filters of different sizes to extract multi-scale feature maps. By fusing these feature maps, the pattern of emotions related to multiple electrode signals and local electrode signals can both be adaptively captured. Finally, the classification results are produced. For more implementation details of network architectures, please refer to the supplemental material.

\subsection{Multi-factor Training Network}




By optimizing the adversarial loss, the trained generator of AAN can produce augmented samples varying in signal details but fitting the feature distribution of real samples. For the next step, we train classifier $C$ and fine-tune the trained classifier further utilizing the learned generator of AAN to generate augmented samples.

In this stage, how to utilize augmented samples for supervision is a crucial problem. It is known that high valence (arousal) and low valence (arousal) are far from being clear-cut and distinguished by an artificial threshold. Thus, if the shift of augmented EEG varies in wide limits, the threshold between high valence (arousal) and low valence (arousal) can be exceeded, and the augmented EEG would change to a different category of the original EEG. To tackle this bottleneck, we explore to seek a self-supervised learning framework to supervise emotion recognition training based on augmented samples and uncertainty labels. It is important to acknowledge that in the field of computer vision, self-supervised learning frameworks already allow for reasonable performance without the acquisition of large training sets and well-labeled training samples. For example, Dosovitskiy \textit{et al.} \cite{dosovitskiy2015discriminative} proposed to learn a network to discriminate between a set of surrogate classes formed by applying various transformations to a randomly sampled ``seed'' image patch. Then, by learning to classify different transformed samples of seed images to the same surrogate categories, the proposed network can extract discriminative features favoring better classification performance.

Thus, inspired by the significant progress made by self-supervised learning, this paper proposes the MTN for EEG-based emotion recognition. As described before, the augmented samples are synchronized based on parts of original EEG signal values, and thus the augmented samples should preserve the feature distribution of original samples to some extent, although not the same. Based on these observations, different from existing work, which directly creates a set of surrogate classes, this paper designs a set of surrogate confidence, measured by augmentation factor $\tau$, learning to restrict the feature distribution difference between real samples and augmented samples under given surrogate confidence. To be specific, in the case where the augmentation factor $\tau$ is large, limited signal values of augmented samples remain from the source EEG signal, and the generator $G$ cannot capture and preserve the feature of the original EEG signal during synchronizing. Thus, the feature distribution of the augmented EEG signals should be constrained to fit the original EEG signals' feature distribution under low confidence. Conversely, if the augmentation factor $\tau$ is small, most original EEG signal values are preserved during augmentation. We should force the feature distribution of augmented samples close to the original samples with high confidence. Finally, we propose Multi-factor Self-supervised Learning loss to assign different weights for restricting the feature distribution difference between augmented EEG signals and real samples based on corresponding surrogate confidence. Combining the cross-entropy loss for supervising real samples' training as usual, and the total loss function can be formulated as:

\begin{equation}
  \label{eq:c_loss}
  \begin{aligned}
  L_{C}(\tau)=&-\frac{1}{n} \sum_{i=0}^{n} y_i \log \left(C(e_i)\right) \\
  +& \frac{\lambda_{a}}{n} \sum_{i=0}^{n} (1-\tau_i) \left\|C_{x}(G(\delta(e_i, \tau_i)))-C_{x}(e_i)\right\|_{2}^{2}
  \end{aligned}
\end{equation}
where we disregard the last fully connected layer of the classifier $C$ and utilize the remaining part as a feature extractor $C_x$ to process EEG signals and produce feature vectors. The ground-truth label of given EEG signal $e_i$ is denoted by $y_i$, $n$ is the number of samples in a mini-batch, and $\lambda_{a}$ is the hyper-parameter utilized to represent the importance of classifying augmented samples correctly.

\begin{figure*}[]
  \begin{center}
      \includegraphics[width=0.85\linewidth]{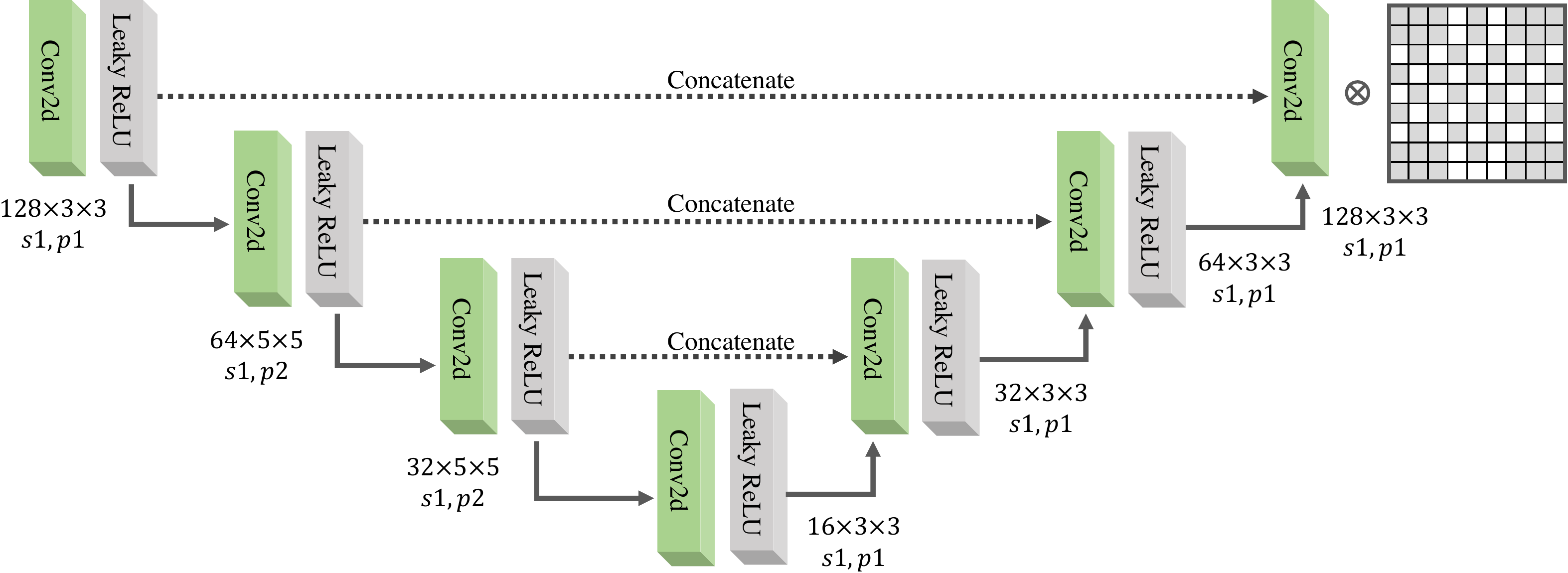}
  \end{center}
  \caption{The network architecture of the generator $G$ in the proposed GANSER framework.}
  \label{supp_g_model}
\end{figure*}

In Eq. (\ref{eq:c_loss}), the cross-entropy between the ground truth labels and corresponding prediction results of the real EEG signals are formulated as $-\frac{1}{n} \sum_{i=0}^{n} y_i \log \left(C(e_i)\right)$. Meanwhile, the feature distribution difference between the original EEG signals and the corresponding augmented signals is denoted by $\left\|C_{x}(G(\delta(e_i, \tau_i)))-C_{x}(e_i)\right\|_{2}^{2}$. Based on augmentation factor $\tau_i$, we compute the surrogate confidence $(1-\tau_i)$ to assign large weights to different augmented samples, to narrow the feature distribution difference when most of the original EEG signal values are preserved in the augmented sample.

It is worth noting that existing data augmentation methods mostly provide augmented samples in an offline fashion, generating a preset number of samples and saving them as training samples for the first step. Then, during optimization, augmented samples are loaded and fed into models with original samples. In this paper, inspired by self-supervised learning approaches, we attempt to explore an alternative strategy. On the one hand, the separated stages of augmentation and training are joined together as an end-to-end pipeline. Given a batch of samples, we first utilize the generator $G$ in AAN to augment EEG signals and pair the real ones. Then, we utilize these samples to optimize the classifier $C$ with Eq. (\ref{eq:c_loss}) in the current batch. Avoiding the cost of saving and reloading, the proposed method is more efficient. On the other hand, in this paper, augmented samples are regenerated between epochs in runtime. Benefit from the randomness of the Masking Transformation operation, augmented samples of the corresponding batch between epochs are different, but both sampled near the real distributions. In this way, instead of overfitting on a preset number of augmented samples, randomly synthesized EEG signals can approximate the distribution of EEG real signals without number limitation.

Finally, the proposed Multi-factor Self-supervised Learning loss can adaptively guide the feature extractor to extract distinguishable feature representation for simulated EEG signals and generalize the classifier to the augmented sample space.

\section{Implementation Details}
In this section, we supplement the implementation details of our proposed framework, GANSER, including the network architectures of the designed generator $G$, discriminator $D$ and classifier $C$, and the hyper-parameters utilized in the experiments.

\begin{figure*}[]
  \begin{center}
      \includegraphics[width=0.85\linewidth]{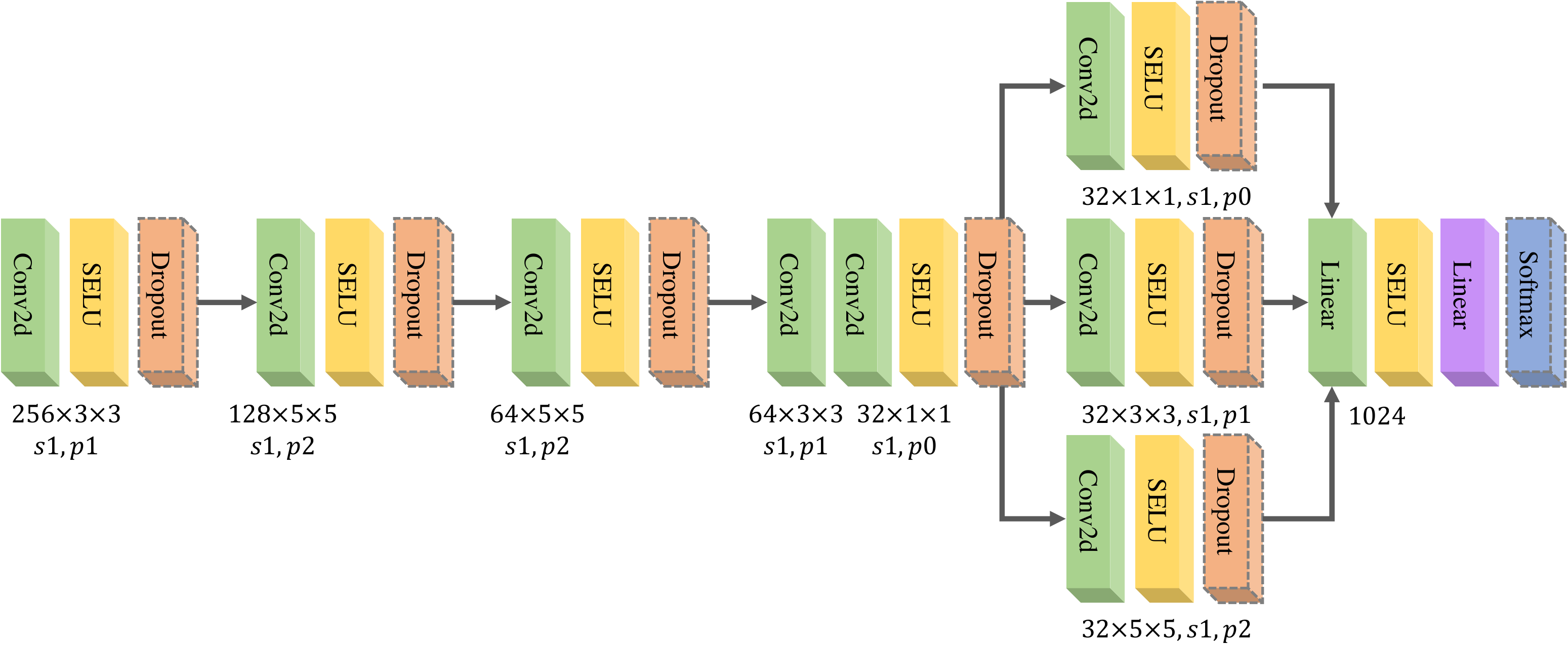}
  \end{center}
  \caption{The network architecture of the STNet utilized by the discriminator $D$ and the classifier $C$ in the proposed GANSER framework. Especially, the dashed box represents operations that exist in the classifier $C$ but do not exist in the discriminator $D$.}
  \label{supp_d_model}
\end{figure*}

\subsection{Network Architectures}
In the proposed GAN, the generator $G$ is trained to generate the simulated EEG sample $G(\delta(e, \tau))$ taking the EEG signal $e$ as input. The network architecture is illustrated in Fig. \ref{supp_g_model}. It consists of a contracting path (left side) and an expansive path (right side). The contracting path follows the typical architecture of an encoder convolutional neural network to extract to down-sample the EEG signals $G(\delta(e, \tau))$ and extract feature maps. It consists of the repeated application of one $3 \times 3$ convolutional layer, two $5 \times 5$ convolutional layer, another $3 \times 3$ convolutional layer, each followed by a leaky rectified linear unit (LeakyReLU). At each convolutional layer, we set the stride to one and halved the number of feature channels. Meanwhile, the expansive path follows a decoder convolutional neural network architecture with skip connections to up-sample feature maps to a high spatio-temporal resolution to generate new EEG samples based on extracted features. Every step in the expansive path consists of concatenation with the correspondingly feature map from the contracting path and one $3 \times 3$ convolutional layer with the stride of one, doubling the number of feature channels. Each convolutional layer is followed by the LeakyReLU activation function. After the final convolutional layer, the Channel Masking operation is introduced to build the channel mask $m$, a prior binary mask with the size of $9 \times 9$, and set the values where electrodes exist and signals are measured to 1.0, while defining the other locations to 0.0. Then, we apply element-wise multiply between the designed prior mask and the output of UNet, i.e., the synthesized EEG signals, to artificially reset the signal values where electrodes do not exist to zero. The total $G$ network has seven convolutional layers.

\begin{table*}[]
  \caption{Hyper parameters utilized in the proposed GANSER framework.}
  \begin{center}
    \begin{tabular}{|l|c|c|c|c|c|c|c|c|}
      \hline
      Stage        & Epoch & $lr_G$  & $lr_D$  & $lr_C$ & $\lambda_p$ & $\lambda_a$ & $\tau_{min}$ & $\tau_{max}$\\ \hline
      AAN          & 300   & 0.00001 & 0.00001 & -        & 1.0       & -           & 0.0        & 0.5       \\ \hline
      MTN          & 300   & -       & -       & 0.00001  & -         & 0.5         & 0.5        & 0.9       \\ \hline
    \end{tabular}
  \end{center}
  \label{hyper_parameter}
  \end{table*}

The discriminator $D$ is required to distinguish whether the given EEG signals are simulated or real. As shown in Fig. \ref{supp_g_model}, in this paper, we design a novel network architecture, STNet, to analyze the complex spatio-temporal features of EEG signals. As shown in Fig. \ref{supp_d_model}, given a real EEG signal or generated EEG signal, the input EEG signal is provided as the input of three convolutional blocks to extract low-resolution features. The first convolutional block uses a $3 \times 3$ convolutional layer, while the second and third convolutional block uses a $5 \times 5$ convolutional layer. Each convolutional layer halves the number of input feature channels, and the stride of all convolutional layers is set to one. Following each convolutional layer, a scaled exponential linear unit (SELU) is utilized as an activation function. For the next step, we introduce a separable convolutional layer \cite{chollet2017xception} to decouple the modeling of spatio-temporal information. In detail, the separable convolutional layer factorizes a standard $3 \times 3$ convolutional layer into a $3 \times 3$ depth-wise convolution and a $1 \times 1$ point-wise convolution and splits the computation into two steps: the depth-wise convolution applies a single convolutional filter per input channel, and the point-wise convolution is used to create a linear combination of the output of the depth-wise convolution. The stride of both depth-wise convolution and point-wise convolution is set to one, and the point-wise convolution is followed by SELU. Moreover, we find recognizing the pattern of emotions requires analyzing EEG signals at different spatial scales, thus introducing an Inception block inspired by \cite{szegedy2015going}. In detail, this paper combines a $1 \times 1$ convolutional layer, a $3 \times 3$ convolutional layer, and a $5 \times 5$ convolutional layer to utilize their output filter banks concatenated into single feature maps forming the input of the next stage. At each convolutional layer, we set the stride to one and halved the number of feature channels. Finally, the feature maps extracted by convolutional layers of each EEG signal are reshaped to a $1\times 2592 (32 \times 9 \times 9)$ feature vector, and two linear layers are utilized to map the feature vector to a scalar. Here, the first linear layer has 1024 nodes and is followed by the SELU activation function, while the second linear layer has one node to predict the score of the given EEG signal being fake.

Because the designed STNet shows good effectiveness in EEG analysis, in this paper, the classifier $C$ shares the design of STNet as the discriminator $D$, except for several modifications to fit the classification problem formulation of emotion recognition. In detail, as shown in Fig. \ref{supp_d_model}, the dashed box represents operations that exist in the classifier $C$ but do not exist in the discriminator $D$. In the classifier $C$, we employ dropout \cite{srivastava2014dropout} operation with $p=0.5$ after the activation function of convolutional layers to address the problem of over-fitting. Here, $p$ denotes the probability of an element to be zeroed. The number of nodes in the last linear layer is set to two or four, corresponding to the categories of emotions. Moreover, an additional softmax function is set to follow the last linear layer as the activation function to produce the final outputs representing a categorical distribution.

\subsection{Other Configuration}

In this paper, we use PyTorch \cite{paszke2019pytorch} to implement our networks based on eight NVIDIA Tesla V100 GPUs. For the networks $G$, $D$ and $C$, this paper adopts the Adam optimizer \cite{kingma2014adam} to minimize the loss functions. Here, the coefficient $\beta_1$ used for computing running averages of the gradient is set to 0.9, and the coefficient $\beta_2$ used for computing running averages of the square is 0.99. Besides, the weight decay of the L2 penalty is set to $0.0005$, and the batch size is 64. Table \ref{hyper_parameter} reports other hyper-parameters utilized in the proposed GANSER. Here, $lr_G$ denotes the learning rate of the generator $G$, $lr_D$ corresponds to the learning rate of the discriminator $D$, and $lr_C$ represents the learning rate of the classifier $D$. In addition, $\lambda_p$ is a hyper-parameter presenting the weight of the penalty term in Eq. (\ref{eq:d_loss}), and $\lambda_{a}$ denotes the hyper-parameter in Eq. (\ref{eq:c_loss}) utilized to represent the importance of classifying augmented samples correctly. The parameter $\tau$ is sampled from the uniform distribution $U\sim[\tau_{min}, \tau_{max}]$. Moreover, in the proposed GANSER framework, the AAN is first trained to generate augmented EEG samples, and MTN is then introduced to train the classifier for the emotion recognition task. Thus, as shown in Table \ref{hyper_parameter}, we report all hyper-parameters utilized in these two stages.

\section{Experiments}
\subsection{Dataset and Preprocessing on DEAP}


To evaluate the proposed method, we conduct experiments on the widely-used EEG-based emotion recognition dataset DEAP \cite{koelstra2011deap} to demonstrate the performance gain of our proposed method. To be specific, the DEAP dataset recorded 32-channel EEG signals and 8-channel peripheral physiological signals of 32 subjects when watching 40 one-minute-long music videos. After watching each video, participants rate their arousal levels, valence, liking, and dominance from one to nine for each video. The EEG signals and rating values are utilized to construct the emotion recognition task.

For EEG signals of each trail, the two preprocessing steps pre-given by the DEAP dataset are first employed. Here, the recorded EEG signals are first down-sampled to a 128 Hz sampling rate. Then, obtained EEG signals are processed with a band-pass filter from 4Hz to 45Hz to remove physiological and power frequency noises \cite{koelstra2011deap}. In each trial of the preprocessed dataset, the contained EEG signals consist of a three-second-long baseline signal recorded in relax state and a 60-second-long experimental signal recorded under stimulation. Further, we use a non-overlapping sliding window to separate the trial data into one-second-long chunks and construct the separated EEG signals as data samples. Here, the sliding window size is set to 128 to separate one-second chunks under a sampling rate of 128 Hz. For the next step, to reduce the effect of basic emotional state, following existing work, we remove a mean baseline value from each epoch \cite{cui2020eeg}. Finally, the total number of EEG samples from 40 trials is  $40\times 60=2400$.

In terms of the emotional rating value of each trail in the range of 1.0 to 9.0 in the arousal and valence domains, the median 5.0 was used as the threshold to divide the rating value into two categories. In the cases where the emotional rating value is rated more than 5.0, the corresponding EEG signals are labeled as high arousal or valence. On the contrary, for the ones less than or equal to 5.0, this paper labels the corresponding EEG signals as low arousal or valence. Finally, given EEG signals to predict corresponding labeled categories, the emotion recognition task is formulated as a binary classification problem.

\subsection{Overall Performance on DEAP}
\label{sec:overall}

In this section, to validate the performance of the proposed framework, we give the overall performance evaluated on the DEAP dataset and compare other state-of-the-art works and competitive GAN-based studies on the emotion recognition task.

In the experiments, data samples in the DEAP dataset are split into five folds at random and five-fold cross-validation is used to evaluate all models. To evaluate our proposed method, we first utilize 80\% randomly shuffled data samples as training data to train the AAN for 300 epochs. In the following step, we fix the AAN parameters and then take the pre-trained AAN to generate augmented samples. Then, the proposed classifier is learned on each fold for 300 epochs, and we supplement the augmented samples generated by AAN for fine-tuning of 300 epochs with the help of MTN. Finally, the fine-tuned models are utilized for evaluation. To assess the overall performance, the average classification accuracies over five folds are reported.

\begin{table}
    \small
    \caption{Average ACC(\%) of GAN-based and other state-of-the-art methods on the DEAP dataset for valence classification, arousal classification and four classification.}
    \begin{center}
      \begin{tabular}{|c|l|c|c|c|}
        \hline
        \multicolumn{2}{|c|}{Method}            & Valence & Arousal & Four \\ \hline
        \multirow{5}{*}{SOTA}      & CDCN \cite{gao2020channel}       & 92.24  & 92.92  & - \\ \cline{2-5} 
                                   & MMResLSTM \cite{ma2019emotion}   & 92.87  & 92.30  & - \\ \cline{2-5} 
                                   & PCRNN \cite{yang2018emotion}     & 90.8   & 91.03  & - \\ \cline{2-5} 
                                   & CNNLSTM \cite{ozdemir2020eeg}    & 90.62  & 86.13  & - \\ \cline{2-5} 
                                   & MergedLSTM \cite{garg2019merged} & 84.89  & 83.85  & - \\ \hline
        \multirow{2}{*}{GAN-based} & MCLFS-GAN \cite{dong2020multi}   & -      & -      & 81.32 \\ \cline{2-5} 
                                   & sWGAN \cite{luo2020data}         & -      & -      & 49.10 \\ \hline
  
        Proposed                   & GANSER     & \textbf{93.52}   & \textbf{94.21} & \textbf{89.74}   \\ \hline
        \end{tabular}
    \end{center}
    \label{overall_exp_deap}
    \vspace{-16pt}
\end{table}

Illustrated in Table \ref{overall_exp_deap}, we first compared our proposed GANSER with five state-of-the-art studies, i.e., CDCN \cite{gao2020channel}, MMResLSTM \cite{ma2019emotion}, PCRNN \cite{yang2018emotion}, CNNLSTM \cite{ozdemir2020eeg}, and MergedLSTM \cite{garg2019merged}, on the DEAP dataset, respect to the emotion dimensions including valence and arousal. These studies develop different network architectures and strategies for emotion recognition. For example, CNNLSTM \cite{ozdemir2020eeg} combined convolutional neural networks and long short-term memory networks to extract distinguishable features, and MMResLSTM \cite{ma2019emotion} further utilized multi-modal information to improve the classification performance. From Table \ref{overall_exp_deap}, we can find the proposed method outperform all state-of-the-art studies on both valence and arousal dimension. Although the designed classifier requires lightweight training parameters and only consists of convolutional layers, the proposed method shows great classification performance of over 93\% for two-dimensional classification tasks and considerably outperforms the second-best method by a margin of near 1.0\%. The comparison results demonstrate the effectiveness of our proposed methods for EEG-based emotion recognition.

Further, to verify the capability of our proposed method in the field of data augmentation for emotion recognition, we further compare the GANSER with several GAN-based data augmentation frameworks. Especially, following the experimental setting of existing GAN-based methods \cite{dong2020multi,luo2020data}, we formulate the emotion recognition task as a four-category classification problem, which aims at distinguishing EEG signals of four categories: high valence and high arousal, high valence and low arousal, low valence and high arousal, and low valence and high arousal. In Table \ref{overall_exp_deap}, it can be found that the proposed GANSER significantly outperforms existing GAN-based data augmentation frameworks with a large margin of over 8.0\%. We can also find that even in the formulation of four-category classification, our proposed method can correctly classify nearly 90\% EEG signals at valence and arousal dimensions simultaneously due to the well-designed AAN and MTN.

\subsection{Ablation Study on DEAP}

\begin{table}
  \small
  \caption{Ablation study on our modules: the Generative Adversarial Network (GAN), the Masking Transformation (MT) operation, and the Multi-factor Self-supervised Learning (MSL) loss. The average accuracies (\%) of different stripped-down versions of our proposed are reported on the DEAP dataset to classify valence and arousal of emotions.}
  \begin{center}
    \begin{tabular}{|c|c|c|c|c|}
      \hline
      \multicolumn{3}{|c|}{Method}                 & \multicolumn{2}{c|}{Metric} \\ \hline
      GAN          & MT & MSL & Valence       & Arousal      \\ \hline
                   &             &                 & 91.39         & 92.22        \\ \hline
      \checkmark   &             &                 & 92.27         & 92.76        \\ \hline
      \checkmark   & \checkmark  &                 & 93.28         & 93.12        \\ \hline
      \checkmark   & \checkmark  & \checkmark      & \textbf{93.52}& \textbf{94.21}        \\ \hline
    \end{tabular}
  \end{center}
  \label{ablation_exp}
  \vspace{-12pt}
\end{table}

In this section, we further conduct an ablation study to investigate the performance gain brought by each key component in our model, including the GAN, the Masking Transformation operation, and the Multi-factor Self-supervised Learning loss (MSL), and provide the performance of stripped-down versions by removing these components one by one.

When the Multi-factor Self-supervised Learning loss is ablated, the augmented samples are treated as an equivalent of given training samples to optimize the cross-entropy based on the label of original EEG signals. After removing the Masking Transformation operation, the proposed generator $G$ takes EEG signal $e$ as input directly instead of the masked EEG signal $\delta(e,\tau)$. While the GAN is further removed, no data augmentation operation is employed, and the classifier is fine-tuned based on given training samples. Finally, we follow the above experimental setting to train and evaluate stripped-down versions of our approach and report the average classification accuracies of different models in Table \ref{ablation_exp}.

It has been shown that our proposed classifier is a strong and effective baseline even in the case where no data augmentation is utilized, and the classifier is fine-tuned based on the original training dataset. The proposed method can achieve the performance of 91.39\% and 92.22\% at valence and arousal dimensions, comparable to the latest arts designed based on complex network architectures or applying multi-modal information. Meanwhile, the design of both GAN and Masking Transformation operation brings a performance gain of about 1.0\% and obviously improves the classification accuracy of EEG signals. This phenomenon demonstrates that with the help of designed components, the framework synthesizes simulated EEG signals to favor learning the EEG signal patterns for emotion recognition. Finally, the additional Multi-factor Self-supervised Learning loss succeeds in enhancing the classification performance to over 93.5\% by further considering the uncertainty of augmented samples' labels and providing further guidance for model training in a self-supervised learning fashion.

\begin{table}
  \small
  \caption{FSTD scores of different generative models are reported. The compared methods include three stripped-down versions of our approach to evaluate the performance gain of the UNet network architecture (UNet), the Channel Masking (CM) operation, and the STNet architecture (STNet).}
  \begin{center}
    \begin{tabular}{|c|c|c|c|c|c|}
      \hline
      \multicolumn{3}{|c|}{Method}                                                                                      & \multirow{2}{*}{Valence} & \multirow{2}{*}{Arousal} & \multirow{2}{*}{Mean} \\ \cline{1-3}
      UNet                            & \multicolumn{1}{l|}{CM} & \multicolumn{1}{l|}{STNet}                     &                          &                          &                       \\ \cline{1-6} 
      \multicolumn{1}{|c|}{}           & \checkmark                   &               \checkmark                         & 96.78          & 80.81                   & 88.80                \\ \cline{1-6} 
      \multicolumn{1}{|c|}{\checkmark} &                              &               \checkmark                         & 103.62              & 72.26                    & 87.94                \\ \cline{1-6} 
      \multicolumn{1}{|c|}{\checkmark} & \checkmark                   &                                                  & 278.76            & 286.63           & 282.70        \\ \hline
      \multicolumn{1}{|c|}{\checkmark} & \checkmark                   &               \checkmark                         & \textbf{49.61}                   & \textbf{36.61}                    & \textbf{43.11}                \\ \hline
    \end{tabular}
  \end{center}
  \label{ablation_exp_fid}
  \vspace{-18pt}
\end{table}

\subsection{Qualitative and Quantitative Experiments on DEAP}

While the overall framework is introduced to improve emotion recognition accuracy, the well-designed network architecture of GAN is especially crucial to synthesize real-like and diverse EEG signals. Thus, in this section, qualitative and quantitative experiments are carried out to assess the quality of the synthesized EEG samples and demonstrate the contribution of our designed GAN architectures for EEG signal stimulation.

In terms of qualitative experiments, it is known that judging the quality and diversity of samples generated by GAN-based methods is a challenging task due to the difficulty of measuring the distribution difference between real samples and generated samples. Following the existing work in the computer vision field, this paper adapts Fréchet Inception Distance (FID) for the EEG analysis and designs Fréchet STNet Distance (FSTD) to assess the quality of EEG signals generated by GANs.

\begin{figure*}[]
  \begin{center}
      \includegraphics[width=\linewidth]{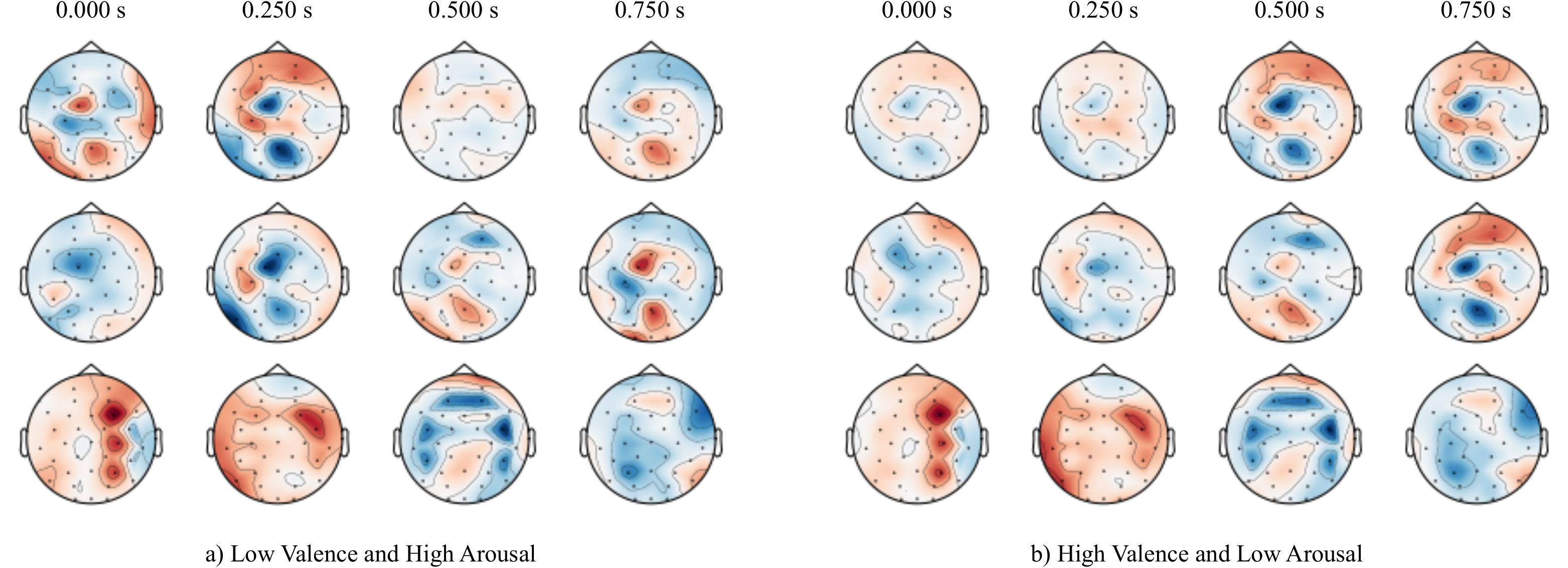}
  \end{center}
  \caption{A few qualitative results are showing the real EEG signals (the first line), augmented EEG signals (the second line) synthesized by the proposed method, and augmented EEG signals (the third line) produced by the proposed method w/o UNet.}
  \label{visual_exp}
\end{figure*}

In detail, we first train two proposed STNet on 80\% training samples to distinguish different emotions at the assessed dimension, e.g., valence dimension. Then, we utilize the learned STNet as the feature extractor and provide EEG samples as input to extract the output of the penultimate fully connected layer. Finally, compute the sample means $\mu$, $\mu_{g}$ and the sample covariance matrices $\Sigma$, $\Sigma_{g}$ of real samples and generated samples' the feature distributions, and the FSTD is then the Wasserstein distance between the two multivariate normal distributions $N(\mu, \Sigma)$ and $N(\mu_{g}, \Sigma_{g})$:

\begin{equation}
  \label{eq:fstd_score}
  \operatorname{FSTD}=\| \mu-\mu_{g}||^{2}+\operatorname{Tr}\left(\Sigma - \Sigma_{g}-2 \sqrt{\Sigma \Sigma_{g}}\right)
\end{equation}
where a small FSTD value indicates a high similarity between the generated samples and real data distribution.

Illustrated in Table \ref{ablation_exp_fid}, FSTD scores of different stripped-down GAN-based architectures of our approach are compared in terms of valence and arousal dimensions. The average FSTD scores are also provided to assess overall performance. Here, to explore the contribution of each design in the network architecture, we use the above experimental setting to train and evaluate different combinations of the UNet network architecture (UNet), the Channel Masking operation, and the STNet architecture (STNet). Each ablated combination removed one of the components. Especially when UNet is removed, we utilize an auto-encoder of the same parameter volume as an alternative following the existing work. After eliminating the Channel Masking operation, the output of UNet is directly utilized as generated EEG signals. While STNet is ablated, we replace the separable convolutional layer and the Inception block with convolutional layers to produce output feature maps of the corresponding size. 

As shown in Table \ref{ablation_exp_fid}, it can be found that if any one of the proposed components is removed, the average FSTD scores will rise. It means that every proposed module contributes to the quality and diversity of augmented samples, without which the feature distribution of the generated sample would be different from the real sample. Especially, we can also find STNet affects the FSTD score to the greatest extent, and in both valence and arousal dimensions. Because favored by the designed separable convolutional layer and the Inception block, STNet can model the complex spatio-temporal feature distribution of EEG signals and force the generated samples to fit the real feature distribution. Meanwhile, the lack of the Channel Masking operation also leads to a relatively sharp increase of FSTD scores at the valence dimension. This phenomenon is because the difference between channels in the EEG signal is crucial for valence recognition, and the Channel Masking operation can reduce the difficulty of synthesizing the difference between channels by giving prior information. Further, arousal analysis is highly related to the signal value scale. In these cases, UNet preserves the modeled scale information to the greatest extent and thus improves the quality of augmented samples. 

To further enhance the revelation of the data augmentation performance of our proposed framework, in this section, we carry out a visualization experiment on the DEAP dataset. To be specific, the experimental setting in this section is consistent with that in Section \ref{sec:overall}. After training is completed, the learned AAN is utilized to synthesize EEG signals based on given data samples. The original EEG signals and the generated simulated EEG signals are provided in Fig. \ref{visual_exp}. 

In detail, EEG signals of different categories are sampled for visualization, including high valence and low arousal (first column) and low valence and high arousal (second column), respectively. For each case, the original real EEG signals are presented in the first row, the second row depicts synthesized EEG signals by our proposed method, and the third row shows the generated EEG signals by an ablated version of our approach replacing the UNet network architecture with auto-encoder. Since the raw data of EEG signals is not conducive to visually representing the EEG signals' characteristics, the topographic maps of EEG signals sampled at 0.0s, 0.25s, 0.5s, and 0.75s are provided where red denotes high energy values, and blue represents low energy values. As shown in Fig. \ref{visual_exp}, it can be found that the activated areas shown by real EEG signals and corresponding simulated EEG signals symmetric by our proposed method are almost consistent at spatial and temporal dimensions, demonstrating the generated data have similar spatio-temporal data distributions as the real data. Meanwhile, we can also observe a slight shift in the activation area and the change in the activation degree in the augmented EEG samples, indicating that our methods could synthesize diverse simulated samples and avoid identity mapping. On the contrary, we found that the EEG signal samples generated by the ablated model preserve limited spatio-temporal features of original EEG signals. It is also difficult to distinguish between different categories of samples, indicating that the GAN may suffer from mode collapse without a specially designed network.

\subsection{Dataset and Preprocessing on DREAMER}

The DREAMER dataset contains EEG data of 23 subjects, which are collected via 14 EEG electrodes from the subjects when they are watching 18 film clips. Each film clip lasts 65 seconds long to 393 seconds long. The average length of film clips is 199 seconds. The data collection begins with a neutral film clip watching to help the subjects return to the neutral emotional state in each new trial of data collection and serve as the baseline signals. As the pre-given preprocessing steps, all the EEG signals are recorded at a sampling rate of 128 Hz and filtered by band-pass Hamming with linear phase FIR filters. The artifact subspace reconstruction (ASR) method is used for artifact removal. After watching a film clip, subjects rate their levels of arousal and valence from 1.0 to 5.0. Finally, there are experimental signals, baseline signals, and labels in the DREAMER dataset. 

For EEG signals of each trail, we use a non-overlapping sliding window to separate the trail data into one-second-long chunks and construct the separated EEG signals as data samples. Here, the sliding window size is set to 128 to separate one-second chunks under a sampling rate of 128 Hz. For the next step, to reduce the effect of basic emotional state, following existing work, we removed a mean baseline value from each epoch \cite{cui2020eeg}. The length of each experimental signal in the DREAMER dataset is different because each film clip lasts from 65 seconds long to 393 seconds long. As a result, we get a different number of EEG samples for each experimental signal of the DREAMER dataset.

In terms of the emotional rating value of each trial, the median 3.0 is used as the threshold to divide the trials according to the levels of Valence and Arousal. That is, the label is low when the rating is less than or equal to 3.0, and the label is high when the rating is greater than 3.0. In this way, the recognition task is actually a binary classification problem for each emotion dimension. 

\subsection{Overall Performance on DREAMER}

In this section, without changing the network architecture and other configurations, we directly implement our proposed method on the DREAMER dataset and report the performance of our proposed method as a baseline. Notably, the contained EEG data in the DREAMER dataset are collected via 14 EEG electrodes, the EEG signals of 14 channels instead of 32 in the DEAP dataset are transformed into $9 \times 9$ maps according to the electrodes' location based on the international 10-20 system. Thus, the representation of EEG signals in the DREAMER dataset is more sparse. Then, experiments are carried out to compare our proposed method with several state-of-the-art works.

In this experiment, following the experimental setting of existing works, data samples in the DREAMER dataset are split into ten folds at random, and ten-fold cross-validation is used to evaluate all models. Notably, such an experimental setting is different from the configuration of existing studies reported on the DEAP dataset, where five-fold cross-validation is generally used. To evaluate our proposed method, we first utilize 80\% randomly shuffled data samples as training data to train the AAN for 300 epochs. In the following step, we fix the AAN parameters and then take the pre-trained AAN to generate augmented samples. Then, the proposed classifier is learned on training folds for 300 epochs, and we supplement the augmented samples generated by AAN for fine-tuning of 300 epochs with the help of MTN. Finally, the fine-tuned models are utilized for evaluation on the corresponding test folds. To assess the overall performance, the average classification accuracies over ten test folds are reported.

\begin{table}
  \small
  \caption{ACC(\%) of GAN-based and other state-of-the-art methods on the DREAMER dataset for valence classification, arousal classification and four classification.}
  \begin{center}
    \begin{tabular}{|c|l|c|c|}
      \hline
      \multicolumn{2}{|c|}{Method}            & Valence & Arousal \\ \hline
      \multirow{5}{*}{SOTA}      & STRNN \cite{zhang2018spatial}       & 70.80  & 80.30  \\ \cline{2-4} 
                                 & PCRNN \cite{yang2018emotion}    & 79.93  & 81.48  \\ \cline{2-4} 
                                 & DT \cite{yang2018continuous}       & 75.53  & 75.74  \\ \cline{2-4} 
                                 & MLP \cite{yang2018continuous}   & 83.64  & 83.71  \\ \cline{2-4} 
                                 & ContCNN \cite{yang2018continuous}     & 84.54   & \textbf{84.84} \\ \hline

      Proposed                   & GANSER     & \textbf{85.28}   & 84.16   \\ \hline
      \end{tabular}
  \end{center}
  \label{overall_exp_dreamer}
  \vspace{-16pt}
\end{table}

Illustrated in Table \ref{overall_exp_dreamer}, we first compared our proposed GANSER with five state-of-the-art studies, including the spatial-temporal recurrent neural network (STRNN) \cite{zhang2018spatial}, the parallel convolutional recurrent neural network (PCRNN) \cite{yang2018emotion}, the decision tree (DT) \cite{yang2018continuous}, the multi-layer perceptron (MLP) \cite{yang2018continuous}, and the continuous CNN (ContCNN) \cite{yang2018continuous}, on the DREAMER dataset, respect to the emotion dimensions including valence and arousal. These studies develop different network architectures and strategies for emotion recognition. For example, Zhang \textit{et al.} \cite{zhang2018spatial} proposed a spatial-temporal recurrent neural network (STRNN) to investigate both spatial and temporal dependencies of EEG signals and achieve the state-of-the-art. In the experiment, the STRNN \cite{zhang2018spatial} is implemented on the DREAMER dataset to see how much the designed RNN architecture can improve the discriminant ability. Further, the parallel convolutional recurrent neural network (PCRNN) reported by \cite{yang2018emotion} is compared, which introduces baseline signals into pre-processing and proposes a hybrid neural network combining CNN and RNN to learn the compositional spatial-temporal feature of EEG signals. In \cite{yang2018continuous}, the authors proposed a 3D representation of the EEG segment to combine features of signals from different frequency bands while preserving spatial information among channels. Then, the performance of the Decision Tree (DT) and the Multi-Layer Perceptron (MLP) is reported to demonstrate the discriminant ability of the proposed EEG representation. Further, the authors \cite{yang2018continuous} introduced the continuous CNN (ContCNN) to utilize the combination of features of multiple bands to complement each other and achieves comparable results. In this paper, the reported performance of DT, MLP, and ContCNN are all considered for the comparison experiment. From Table \ref{overall_exp_dreamer}, we can find the proposed method outperforms most state-of-the-art studies on both valence and arousal dimensions. Especially, although the designed classifier requires lightweight training parameters and only consists of convolutional layers, the proposed method shows great classification performance of about 85\% for two-dimensional classification tasks and considerably outperforms all the state-of-the-art methods at valence dimension.

\section{Conclusion and Future Work}

In this paper, we propose a novel Generative Adversarial Network-based Self-supervised Data Augmentation (GANSER) framework, consisting of Adversarial Augmentation Network (AAN) and Multi-factor Training Network (MTN). In the proposed framework, we are first to combine adversarial training with self-supervised learning to tackle the EEG-based emotion recognition task. The design of AAN employs the Masking Transformation operation to mask parts of EEG signals and then force a well-designed GAN to generate high-quality and high-diversity simulated EEG samples. Here, simple but effective network architectures, e.g., the UNet with Channel Masking operation and STNet, are employed to capture the complex spatio-temporal features of EEG signals. Further, to effectively utilize simulated EEG signals, we introduce MTN, where the Multi-factor Self-supervised Learning loss is proposed to utilize the masking possibility of the Masking Transformation operation as prior knowledge and guide the feature extraction process of simulated EEG signals for generalizing the classifier to the augmented sample space.

By applying the designed framework on the benchmark datasets for emotion recognition, DEAP, and DREAMER, the experimental results show that the designed model can exploit the natural feature of real EEG signals to synthesize high-quality and diverse simulated EEG signals and finally improve the classification performance. In the future, we will seek to explore a semantic data augmentation framework in which the influence of the environment noise, artifacts, and the feature at valence or arousal dimensions of simulated EEG signals can be further controlled and modified.



\bibliographystyle{IEEEtran}
\bibliography{main}

\begin{thebibliography}{10}
\providecommand{\url}[1]{#1}
\csname url@samestyle\endcsname
\providecommand{\newblock}{\relax}
\providecommand{\bibinfo}[2]{#2}
\providecommand{\BIBentrySTDinterwordspacing}{\spaceskip=0pt\relax}
\providecommand{\BIBentryALTinterwordstretchfactor}{4}
\providecommand{\BIBentryALTinterwordspacing}{\spaceskip=\fontdimen2\font plus
\BIBentryALTinterwordstretchfactor\fontdimen3\font minus
  \fontdimen4\font\relax}
\providecommand{\BIBforeignlanguage}[2]{{%
\expandafter\ifx\csname l@#1\endcsname\relax
\typeout{** WARNING: IEEEtran.bst: No hyphenation pattern has been}%
\typeout{** loaded for the language `#1'. Using the pattern for}%
\typeout{** the default language instead.}%
\else
\language=\csname l@#1\endcsname
\fi
#2}}
\providecommand{\BIBdecl}{\relax}
\BIBdecl

\bibitem{cambria2017affective}
E.~Cambria, D.~Das, S.~Bandyopadhyay, and A.~Feraco, ``Affective computing and
  sentiment analysis,'' in \emph{A Practical Guide to Sentiment Analysis},
  2017, pp. 1--10.

\bibitem{minsky2007emotion}
M.~Minsky, \emph{The emotion machine: Commonsense thinking, artificial
  intelligence, and the future of the human mind}, 2007.

\bibitem{zhong2020eeg}
P.~Zhong, D.~Wang, and C.~Miao, ``{EEG}-based emotion recognition using
  regularized graph neural networks,'' \emph{Trans. Affective Computing}, 2020.

\bibitem{roy2019deep}
Y.~Roy, H.~Banville, I.~Albuquerque, A.~Gramfort, T.~H. Falk, and J.~Faubert,
  ``Deep learning-based electroencephalography analysis: a systematic review,''
  \emph{Journal of Neural Engineering}, vol.~16, no.~5, p. 051001, 2019.

\bibitem{lashgari2020data}
E.~Lashgari, D.~Liang, and U.~Maoz, ``Data augmentation for deep-learning-based
  electroencephalography,'' \emph{Journal of Neuroscience Methods}, p. 108885,
  2020.

\bibitem{fawzi2016adaptive}
A.~Fawzi, H.~Samulowitz, D.~Turaga, and P.~Frossard, ``Adaptive data
  augmentation for image classification,'' in \emph{ICIP}, 2016, pp.
  3688--3692.

\bibitem{8320798}
T.~Song, W.~Zheng, P.~Song, and Z.~Cui, ``{EEG} emotion recognition using
  dynamical graph convolutional neural networks,'' \emph{Trans. Affective
  Computing}, vol.~11, no.~3, pp. 532--541, 2020.

\bibitem{9154557}
X.~Du, C.~Ma, G.~Zhang, J.~Li, Y.-K. Lai, G.~Zhao, X.~Deng, Y.-J. Liu, and
  H.~Wang, ``An efficient {LSTM} network for emotion recognition from
  multichannel {EEG} signals,'' \emph{Trans. Affective Computing}, pp. 1--1,
  2020.

\bibitem{9204431}
W.~Tao, C.~Li, R.~Song, J.~Cheng, Y.~Liu, F.~Wan, and X.~Chen, ``{EEG}-based
  emotion recognition via channel-wise attention and self attention,''
  \emph{Trans. Affective Computing}, pp. 1--1, 2020.

\bibitem{8089737}
H.~Becker, J.~Fleureau, P.~Guillotel, F.~Wendling, I.~Merlet, and L.~Albera,
  ``Emotion recognition based on high-resolution {EEG} recordings and
  reconstructed brain sources,'' \emph{Trans. Affective Computing}, vol.~11,
  no.~2, pp. 244--257, 2020.

\bibitem{9321519}
G.~Zhang, M.~Yu, Y.-J. Liu, G.~Zhao, D.~Zhang, and W.~Zheng, ``{SparseDGCNN}:
  Recognizing emotion from multichannel {EEG} signals,'' \emph{Trans. Affective
  Computing}, pp. 1--1, 2021.

\bibitem{zhang2018spatial}
T.~Zhang, W.~Zheng, Z.~Cui, Y.~Zong, and Y.~Li, ``Spatial--temporal recurrent
  neural network for emotion recognition,'' \emph{Trans. Cybernetics}, vol.~49,
  no.~3, pp. 839--847, 2018.

\bibitem{li2016emotion}
X.~Li, D.~Song, P.~Zhang, G.~Yu, Y.~Hou, and B.~Hu, ``Emotion recognition from
  multi-channel {EEG} data through convolutional recurrent neural network,'' in
  \emph{BIBM}, 2016, pp. 352--359.

\bibitem{salama2018eeg}
E.~S. Salama, R.~A. El-Khoribi, M.~E. Shoman, and M.~A.~W. Shalaby,
  ``{EEG}-based emotion recognition using 3d convolutional neural networks,''
  \emph{IJACSA}, vol.~9, no.~8, pp. 329--337, 2018.

\bibitem{moon2018convolutional}
S.-E. Moon, S.~Jang, and J.-S. Lee, ``Convolutional neural network approach for
  eeg-based emotion recognition using brain connectivity and its spatial
  information,'' in \emph{ICASSP}, 2018, pp. 2556--2560.

\bibitem{moon2020madenet}
S.-E. Moon and J.-S. Lee, ``Madenet: Disentangling individuality of {EEG}
  signals through feature space mapping and detachment,'' in \emph{EMBC}, 2020,
  pp. 260--263.

\bibitem{luo2018wgan}
Y.~Luo, S.-Y. Zhang, W.-L. Zheng, and B.-L. Lu, ``Wgan domain adaptation for
  {EEG}-based emotion recognition,'' in \emph{ICNIP}, 2018, pp. 275--286.

\bibitem{krell2017rotational}
M.~M. Krell and S.~K. Kim, ``Rotational data augmentation for
  electroencephalographic data,'' in \emph{EMBC}, 2017, pp. 471--474.

\bibitem{lotte2015signal}
F.~Lotte, ``Signal processing approaches to minimize or suppress calibration
  time in oscillatory activity-based brain--computer interfaces,''
  \emph{PIEEE}, vol. 103, no.~6, pp. 871--890, 2015.

\bibitem{wang2018data}
F.~Wang, S.-h. Zhong, J.~Peng, J.~Jiang, and Y.~Liu, ``Data augmentation for
  {EEG}-based emotion recognition with deep convolutional neural networks,'' in
  \emph{MM}, 2018, pp. 82--93.

\bibitem{luo2020data}
Y.~Luo, L.-Z. Zhu, Z.-Y. Wan, and B.-L. Lu, ``Data augmentation for enhancing
  {EEG}-based emotion recognition with deep generative models,'' \emph{Journal
  of Neural Engineering}, vol.~17, no.~5, p. 056021, 2020.

\bibitem{luo2018eeg}
Y.~Luo and B.-L. Lu, ``{EEG} data augmentation for emotion recognition using a
  conditional {Wasserstein GAN},'' in \emph{EMBC}, 2018, pp. 2535--2538.

\bibitem{luo2019gan}
Y.~Luo, L.-Z. Zhu, and B.-L. Lu, ``A {GAN}-based data augmentation method for
  multimodal emotion recognition,'' in \emph{ISNN}, 2019, pp. 141--150.

\bibitem{corley2018deep}
I.~A. Corley and Y.~Huang, ``Deep {EEG} super-resolution: Upsampling {EEG}
  spatial resolution with generative adversarial networks,'' in \emph{BHI},
  2018, pp. 100--103.

\bibitem{ma2019depersonalized}
B.-Q. Ma, H.~Li, Y.~Luo, and B.-L. Lu, ``Depersonalized cross-subject vigilance
  estimation with adversarial domain generalization,'' in \emph{IJCNN}, 2019,
  pp. 1--8.

\bibitem{aznan2019simulating}
N.~K.~N. Aznan, A.~Atapour-Abarghouei, S.~Bonner, J.~D. Connolly,
  N.~Al~Moubayed, and T.~P. Breckon, ``Simulating brain signals: Creating
  synthetic {EEG} data via neural-based generative models for improved ssvep
  classification,'' in \emph{IJCNN}, 2019, pp. 1--8.

\bibitem{luo2020eeg}
T.-j. Luo, Y.~Fan, L.~Chen, G.~Guo, and C.~Zhou, ``{EEG} signal reconstruction
  using a generative adversarial network with wasserstein distance and
  temporal-spatial-frequency loss,'' \emph{Frontiers in Neuroinformatics},
  vol.~14, 2020.

\bibitem{roy2020mieeg}
S.~Roy, S.~Dora, K.~McCreadie, and G.~Prasad, ``{MIEEG-GAN}: Generating
  artificial motor imagery electroencephalography signals,'' in \emph{IJCNN},
  2020, pp. 1--8.

\bibitem{yang2018emotion}
Y.~Yang, Q.~Wu, M.~Qiu, Y.~Wang, and X.~Chen, ``Emotion recognition from
  multi-channel {EEG} through parallel convolutional recurrent neural
  network,'' in \emph{IJCNN}, 2018, pp. 1--7.

\bibitem{NIPS2017_7159}
I.~Gulrajani, F.~Ahmed, M.~Arjovsky, V.~Dumoulin, and A.~C. Courville,
  ``Improved training of {Wasserstein GANs},'' in \emph{NIPS}, 2017, pp.
  5767--5777.

\bibitem{chollet2017xception}
F.~Chollet, ``Xception: Deep learning with depthwise separable convolutions,''
  in \emph{CVPR}, 2017, pp. 1251--1258.

\bibitem{szegedy2015going}
C.~Szegedy, W.~Liu, Y.~Jia, P.~Sermanet, S.~Reed, D.~Anguelov, D.~Erhan,
  V.~Vanhoucke, and A.~Rabinovich, ``Going deeper with convolutions,'' in
  \emph{CVPR}, 2015, pp. 1--9.

\bibitem{dosovitskiy2015discriminative}
A.~Dosovitskiy, P.~Fischer, J.~T. Springenberg, M.~Riedmiller, and T.~Brox,
  ``Discriminative unsupervised feature learning with exemplar convolutional
  neural networks,'' \emph{TPAMI}, vol.~38, no.~9, pp. 1734--1747, 2015.

\bibitem{srivastava2014dropout}
N.~Srivastava, G.~Hinton, A.~Krizhevsky, I.~Sutskever, and R.~Salakhutdinov,
  ``Dropout: A simple way to prevent neural networks from overfitting,''
  \emph{The Journal of Machine Learning Research}, vol.~15, no.~1, pp.
  1929--1958, 2014.

\bibitem{paszke2019pytorch}
A.~Paszke, S.~Gross, F.~Massa, A.~Lerer, J.~Bradbury, G.~Chanan, T.~Killeen,
  Z.~Lin, N.~Gimelshein, L.~Antiga \emph{et~al.}, ``{PyTorch}: An imperative
  style, high-performance deep learning library,'' in \emph{NIPS}, 2019, pp.
  8026--8037.

\bibitem{kingma2014adam}
D.~P. Kingma and J.~Ba, ``Adam: A method for stochastic optimization,''
  \emph{CoRR}, 2014.

\bibitem{koelstra2011deap}
S.~Koelstra, C.~Muhl, M.~Soleymani, J.-S. Lee, A.~Yazdani, T.~Ebrahimi, T.~Pun,
  A.~Nijholt, and I.~Patras, ``{DEAP}: A database for emotion analysis; using
  physiological signals,'' \emph{Trans. Affective Computing}, vol.~3, no.~1,
  pp. 18--31, 2011.

\bibitem{cui2020eeg}
H.~Cui, A.~Liu, X.~Zhang, X.~Chen, K.~Wang, and X.~Chen, ``{EEG}-based emotion
  recognition using an end-to-end regional-asymmetric convolutional neural
  network,'' \emph{KBS}, vol. 205, p. 106243, 2020.

\bibitem{gao2020channel}
Z.~Gao, X.~Wang, Y.~Yang, Y.~Li, K.~Ma, and G.~Chen, ``A channel-fused dense
  convolutional network for {EEG}-based emotion recognition,'' \emph{Trans.
  Cognitive and Developmental Systems}, 2020.

\bibitem{ma2019emotion}
J.~Ma, H.~Tang, W.-L. Zheng, and B.-L. Lu, ``Emotion recognition using
  multimodal residual lstm network,'' in \emph{ACMMM}, 2019, pp. 176--183.

\bibitem{ozdemir2020eeg}
M.~A. Ozdemir, M.~Degirmenci, E.~Izci, and A.~Akan, ``{EEG}-based emotion
  recognition with deep convolutional neural networks,'' \emph{Biomedical
  Engineering}, vol.~1, 2020.

\bibitem{garg2019merged}
A.~Garg, A.~Kapoor, A.~K. Bedi, and R.~K. Sunkaria, ``Merged lstm model for
  emotion classification using {EEG} signals,'' in \emph{ICDSE}, 2019, pp.
  139--143.

\bibitem{dong2020multi}
Y.~Dong and F.~Ren, ``Multi-reservoirs eeg signal feature sensing and
  recognition method based on generative adversarial networks,'' \emph{Computer
  Communications}, vol. 164, pp. 177--184, 2020.

\bibitem{yang2018continuous}
Y.~Yang, Q.~Wu, Y.~Fu, and X.~Chen, ``Continuous convolutional neural network
  with 3d input for {EEG}-based emotion recognition,'' in \emph{ICNIP}, 2018,
  pp. 433--443.

\end{thebibliography}
\end{document}